\documentclass[10pt,twocolumn,letterpaper]{article}

\usepackage{cvpr}
\usepackage{times}
\usepackage{epsfig}
\usepackage{graphicx}
\usepackage{amsmath}
\usepackage{amssymb}

\usepackage{booktabs}
\usepackage{caption}
\usepackage{subfig}
\usepackage{bm}

\usepackage{xcolor}
\usepackage{flushend}
\usepackage{threeparttable}

\usepackage[pagebackref=true,breaklinks=true,letterpaper=true,colorlinks,bookmarks=false]{hyperref}

\newcommand{\xs}[1]{\bm{x}_s^\text{#1}}
\newcommand{\ys}{\bm{y}_s^\text{3D}}
\newcommand{\xt}[1]{\bm{x}_t^\text{#1}}
 
\cvprfinalcopy 

\begin{document}

\title{xMUDA: Cross-Modal Unsupervised Domain Adaptation\\ for 3D Semantic Segmentation}

\author{
Maximilian Jaritz\textsuperscript{1,2,3}, 
Tuan-Hung Vu\textsuperscript{3}, 
Raoul de Charette\textsuperscript{1},
\'Emilie Wirbel\textsuperscript{2,3}, 
Patrick P{\'e}rez\textsuperscript{3}\\[1ex]
\textsuperscript{1}Inria, 
\textsuperscript{2}Valeo DAR,
\textsuperscript{3}Valeo.ai
}

\twocolumn[{
	\renewcommand\twocolumn[1][]{#1}
	\maketitle
	\centering
	\vspace{-0.3cm}
	\includegraphics[width=0.9\linewidth]{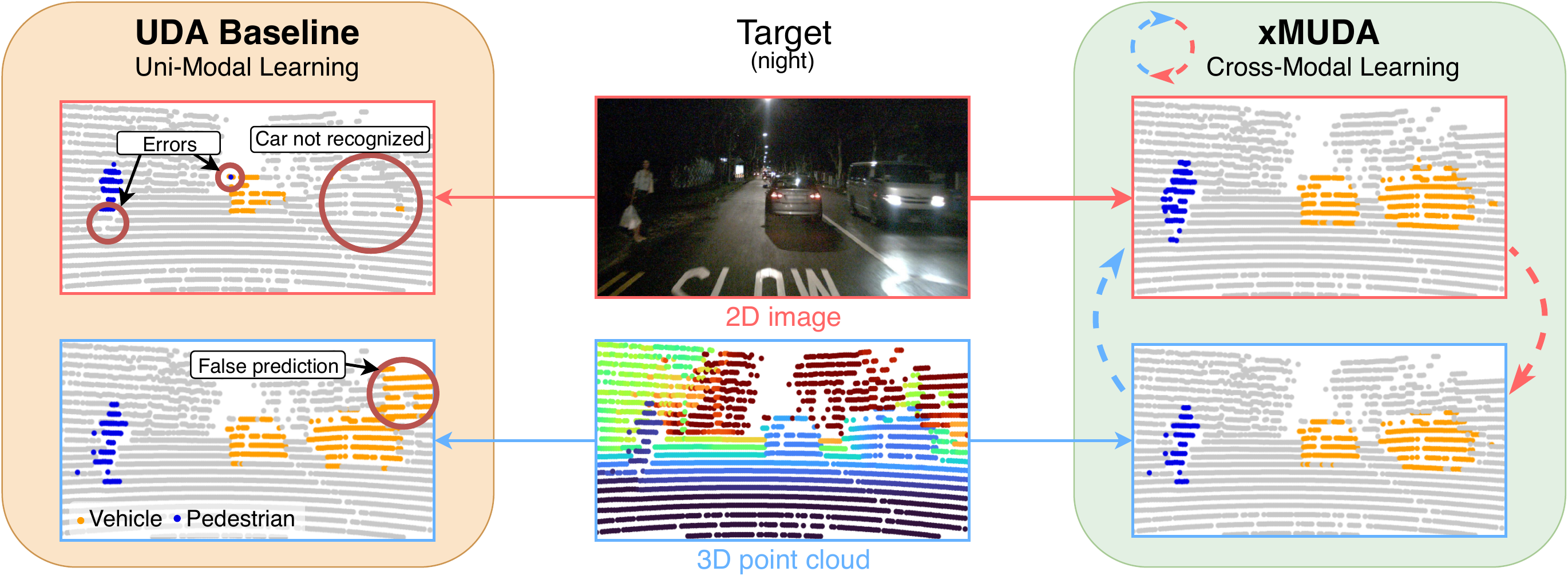}\vspace{-0.3cm}
	\captionof{figure}
	{\textbf{Advantages of cross-modal UDA (xMUDA) in presence of domain gap (day-to-night)}. On this 3D semantic segmentation example, the UDA Baseline~\cite{li2019bidirectional} prediction from {\color{red}2D camera image} does not detect the car on the right due to the day/night domain shift. With xMUDA, 2D learns the appearance of cars in the dark from information exchange with the {\color{cyan} 3D LiDAR point cloud}, and 3D learns to reduce false predictions.
	}
	\label{fig:teaser}
    \vspace{0.4cm}
}]


\begin{abstract}
Unsupervised Domain Adaptation (UDA) is crucial to tackle the lack of annotations in a new domain. There are many multi-modal datasets, but most UDA approaches are uni-modal. 
In this work, we explore how to learn from multi-modality and propose cross-modal UDA (xMUDA) where we assume the presence of 2D images and 3D point clouds for 3D semantic segmentation. 
This is challenging as the two input spaces are heterogeneous and can be impacted differently by domain shift. 
In xMUDA, modalities learn from each other through mutual mimicking, disentangled from the segmentation objective, to prevent the stronger modality from adopting false predictions from the weaker one. 
We evaluate on new UDA scenarios including day-to-night, country-to-country and dataset-to-dataset, leveraging recent autonomous driving datasets. 
xMUDA brings large improvements over uni-modal UDA on all tested scenarios, and is complementary to state-of-the-art UDA techniques.
Code is available at~\emph{\url{https://github.com/valeoai/xmuda}}.
\end{abstract}

\section{Introduction}
Three-dimensional scene understanding is required in numerous applications, in particular robotics, autonomous driving and virtual reality. 
Among the different tasks under concern, 3D semantic segmentation is gaining more and more traction as new datasets are being released~\cite{behley2019iccv, dai2017scannet, aev2019}. 
Like other perception tasks, 3D semantic segmentation can encounter the problem of domain shift between supervised training and test time, for example between day and night, different countries or datasets.
Domain adaptation aims at addressing this gap, but existing work concerns mostly 2D semantic segmentation~\cite{hoffman2018cycada,li2019bidirectional,vu2019advent,zou2019confidence} and rarely 3D~\cite{wu2019squeezesegv2}. We also observe that previous domain adaptation work focuses on single modality, whereas 3D datasets are often multi-modal, consisting of 3D point clouds \textit{and} 2D images.
While the complementarity between these two modalities is already exploited by both human annotators and learned models to localize objects in 3D scenes~\cite{liang2018contfuse, Qi_2018frustumpointnets}, we consider it through a new angle, asking the question:
If 3D and 2D data are available in the source \textit{and} target domain, can we capitalize on multi-modality to address Unsupervised Domain Adaptation (UDA)?

We coin our method \textit{cross-modal UDA}, `xMUDA' in short, and consider 3 real-to-real adaptation scenarios with different lighting conditions (day-to-night), environments (country-to-country) and sensor setup (dataset-to-dataset). It is a challenging task for various reasons. The heterogeneous input spaces (2D and 3D) make the pipeline complex as it implies to work with heterogeneous network architectures and 2D-3D projections. In fusion, if two sensors register the same scene, there is shared information between both, but each sensor also has private (or exclusive) information. Thanks to the latter, one modality can be stronger than the other in a certain case, but it can be the other way around in another, depending on class, context, resolution, etc. This makes selecting the ``best'' sensor based on prior knowledge unfeasible. Additionally, each modality can be affected differently by the domain shift. For example, camera is deeply impacted by the day-to-night domain change, while LiDAR is relatively robust to it, as shown in Fig.\,\ref{fig:teaser}.

In order to address these challenges, we propose a cross-modal UDA (`xMUDA') framework where information can be exchanged between 2D and 3D in order to learn from each other for UDA (see right side of Fig.\,\ref{fig:teaser}). We use a disentangled 2-stream architecture to address the domain gap individually in each modality. Our learning scheme allows robust balancing of the cross-modal and segmentation objective. In addition, xMUDA can be combined with existing \textit{uni-modal} UDA techniques. In this work, we showcase complementarity to self-training with pseudo-labels. Finally, it is common practice in supervised learning to use feature fusion (e.g., early or late fusion) when multiple modalities are available~\cite{hazirbas2016fusenet, liang2018contfuse, Valada_2019}: our framework can be extended to fusion while maintaining a disentangled cross-modal objective.

Our contributions can be summarized as follows:
\begin{itemize}
    \item We define new UDA scenarios and propose corresponding splits on recently published 2D-3D datasets.
    \item We design an architecture that enables cross-modal learning by disentangling private and shared information in 2D and 3D.
    \item We propose a novel UDA learning scheme where modalities can learn from each other in balance with the main objective. It can be applied on top of state-of-the-art self-training techniques to boost performance.
    \item We showcase how our framework can be extended to late fusion and produce superior results.
\end{itemize}
On the different proposed benchmarks we outperform the single-modality state-of-the-art UDA techniques by a significant margin. Thereby, we show that the exploitation of multi-modality for UDA is a powerful tool that can benefit a wide range of multi-sensor applications.

\begin{figure*}
	\centering
	\includegraphics[width=0.95\textwidth]{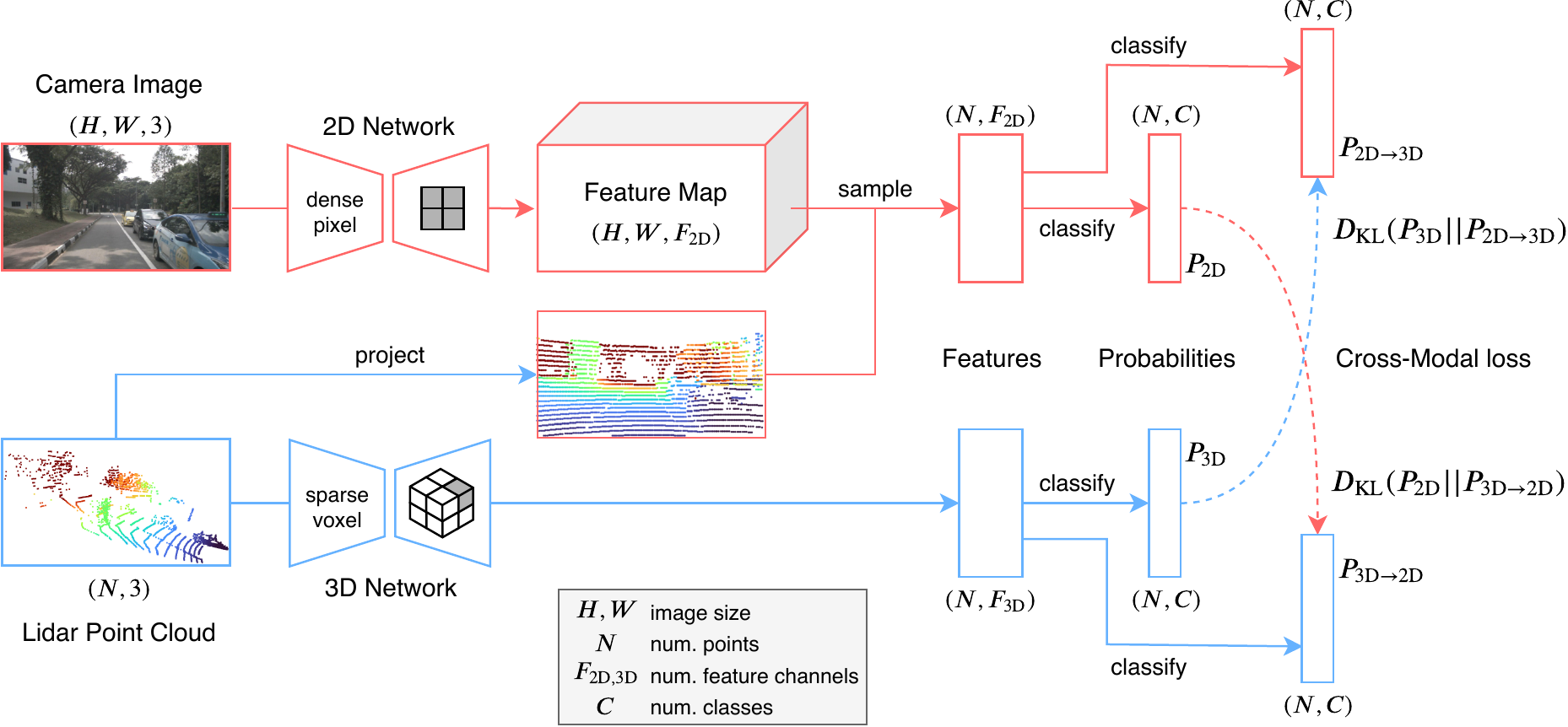}
	\caption{\textbf{Overview of our xMUDA framework for 3D semantic segmentation}. The architecture comprises a 2D stream which takes an image as input and uses a U-Net-style 2D ConvNet~\cite{ronneberger2015unet}, and a 3D stream which takes the point cloud as input and uses a U-Net-Style 3D \mbox{SparseConvNet}~\cite{SparseConvNet}. Feature outputs of both streams have same length $N$, equal to the number of 3D points. To achieve that, we project the 3D points into the image and sample the 2D features at the corresponding pixel locations. The 4 segmentation outputs consist of the main predictions $P_\text{2D}, P_\text{3D}$ and the mimicry predictions $P_{\text{2D} \to \text{3D}}, P_{\text{3D} \to \text{2D}}$. We transfer knowledge across modalities using KL divergence, $D_{\text{KL}}(P_\text{3D} || P_{\text{2D} \to \text{3D}})$, where the objective of the 2D mimicry head is to estimate the main 3D output and vice versa, $D_{\text{KL}}(P_\text{2D} || P_{\text{3D} \to \text{2D}})$.}
	\label{fig:architecture}
	\vspace{-0.3cm}
\end{figure*}

\section{Related Work}

In this section, rather than thoroughly going through the literature, we review representative works for each focus.

\vspace{-0.2cm}
\paragraph{Unsupervised Domain Adaptation.}
The past few years have seen an increasing interest in unsupervised domain adaptation techniques for complex perception tasks like object detection and semantic segmentation.
Under the hood of such methods lies the same spirit of learning domain-invariant representations, i.e., features coming from different domains should introduce insignificant discrepancy.
Some works promote adversarial training to minimize source-target distribution shift, either on pixel-~\cite{hoffman2018cycada}, feature-~\cite{hoffman-arxiv2016} or output-space~\cite{tsai2018learning,vu2019advent}.
Revisited from semi-supervised learning~\cite{lee2013pseudo}, self-training with pseudo-labels has also been recently proven effective for UDA~\cite{li2019bidirectional,zou2019confidence}.

While most existing works consider UDA in the 2D world, very few tackle the 3D counterpart.
Wu~\textit{et al.}~\cite{wu2019squeezesegv2} adopted activation correlation alignment~\cite{morerio2017minimal} for UDA in 3D segmentation from LiDAR point clouds.
In this work, we investigate the same task, but differently: our system operates on multi-modal input data, i.e., RGB + LiDAR.

To the best of our knowledge, there are no previous UDA works in 2D/3D semantic segmentation for multi-modal scenarios.
Only some consider the extra modality, e.g. depth, solely available at training time on source domain and leverage such~\emph{privileged information} to boost adaptation performance~\cite{lee2018spigan,vu2019dada}.
Otherwise, we here assume all modalities are available at train and test time on both source and target domains.

\vspace{-0.2cm}
\paragraph{Multi-Modality Learning.}
In a supervised setting, performance can naturally be improved by fusing features from multiple sources. The geometrically simplest case is RGB-Depth fusion with dense pixel-to-pixel correspondence for 2D segmentation~\cite{hazirbas2016fusenet, Valada_2019}. It is harder to fuse a 3D point cloud with a 2D image, because they live in different metric spaces. One solution is to project 2D and 3D features into a `bird eye view'~\cite{liang2019multi,liang2018contfuse} or `LiDAR front view'~\cite{meyer2019sensor} for the task of object detection. Another possibility is to lift 2D features from multi-view images to the 3D point cloud to enable joint 2D-3D processing for 3D semantic segmentation~\cite{chiang2019unified, Jaritz_2019_ICCV, su2018splatnet}. We are closer to the last series of works: we share the same goal of 3D semantic segmentation. However, we focus on how to exploit multi-modality for UDA instead of supervised learning and only use single view images and their corresponding point clouds.

\vspace{-0.2cm}
\paragraph{3D networks for semantic segmentation.}
While images are dense tensors, 3D point clouds can be represented in multiple ways which leads to competing network families that evolve in parallel. Voxels are very similar to pixels, but very memory intense as most of them are empty. Graham~\textit{et al.}~\cite{SparseConvNet} and similar implementation~\cite{choy20194d} address this problem by using hash tables to convolve only on active voxels. This allows for very high resolution with typically only one point per voxel.
Point-based networks perform computation in continuous 3D space and can thus directly accept point clouds as input. PointNet++~\cite{qi2017pointnetplusplus} uses point-wise convolution, max-pooling to compute global features and local neighborhood aggregation for hierarchical learning akin to CNNs. Many improvements have been proposed in this direction, such as continuous convolutions~\cite{wang2018deepcontinuous} and deformable kernels~\cite{thomas2019kpconv}.
Graph-based networks convolve on the edges of a point point cloud~\cite{Wang_2019dgcnn}.
In this work, we select SparseConvNet~\cite{SparseConvNet} as 3D network which is the state-of-the-art on the ScanNet benchmark~\cite{dai2017scannet}.

\section{xMUDA}

The aim of cross-modal UDA (xMUDA) is to exploit multi-modality by enabling controlled information exchange between modalities so that they can learn from each other. This is achieved through letting them mutually mimic each other's outputs, so that they can both benefit from their counterpart's strengths.

Specifically, we investigate xMUDA using point cloud (3D modality) and image (2D modality) on the task of 3D semantic segmentation. An overview is depicted in Fig.\,\ref{fig:architecture}. We first describe the architecture in Sec.\,\ref{sec:CMLNetwork}, our learning scheme in Sec.\,\ref{sec:CMLLearningScheme}, and later showcase its extension to the special case of fusion.

In the following, we consider a \textit{source} dataset $\mathcal{S}$, where each sample consists of 2D image $\xs{2D}$, 3D point cloud $\xs{3D}$ and 3D segmentation labels $\ys$ as well as a \textit{target} dataset $\mathcal{T}$, lacking annotations, where each sample only consists of image $\xt{2D}$ and point cloud $\xt{3D}$.
Images $\bm{x}^\text{2D}$ are of spatial size $(H, W, 3)$ and point clouds $\bm{x}^\text{3D}$ of spatial size $(N, 3)$, with $N$ the number of 3D points in the camera field of view.

\subsection{Architecture}
\label{sec:CMLNetwork}

To allow cross-modal learning, it is crucial to extract features specific to each modality. 
Opposed to 2D-3D architectures where 2D features are lifted to 3D~\cite{liang2018contfuse}, we use a 2-stream architecture with independent 2D and 3D branches that do \textit{not} share features (see Fig.\,\ref{fig:architecture}).

We use SparseConvNet~\cite{SparseConvNet} for 3D and a modified version of U-Net~\cite{ronneberger2015unet} with ResNet34~\cite{he2016resnet} for 2D. 
Even though each stream has a specific network architecture, it is important that the outputs are of same size to allow cross-modal learning.
Implementation details are provided in Sec.\,\ref{sec:implementation}.

\vspace{-0.2cm}
\paragraph{Dual Segmentation Head.}
We call segmentation head the last linear layer in the network that transforms the output features into logits followed by a softmax function to produce the class probabilities. 
For xMUDA, we establish a link between 2D and 3D with a `mimicry' loss between the output probabilities, i.e., each modality should predict the other modality's output. 
This allows us to explicitly control the cross-modal learning.

In a naive approach, each modality has a single segmentation head and a cross-modal optimization objective aligns the outputs of both modalities. Unfortunately, this leads to only using information that is shared between the two modalities, while discarding private information that is exclusive to each sensor (more details in the ablation study in Sec.\,\ref{sec:segmentationHeads}).
This is an important limitation, as we want to leverage both private and shared information, in order to obtain the best possible performance. 

To preserve private information while benefiting from shared knowledge, we introduce an additional segmentation head to uncouple the mimicry objective from the main segmentation objective. This means that the 2D and 3D streams both have two segmentation heads: one main head for the best possible prediction, and one mimicry head to estimate the other modality's output.

The outputs of the 4 segmentation heads (see Fig.\,\ref{fig:architecture}) are of size $(N, C)$, where $C$ is equal to the number of classes such that we obtain a vector of class probabilities for each 3D point. The two main heads produce the best possible predictions, $P_\text{2D}$ and $P_\text{3D}$ respectively for each branch. The two mimicry heads estimate the other modality's output: 2D estimates 3D ($P_{\text{2D} \to \text{3D}}$) and 3D estimates 2D ($P_{\text{3D} \to \text{2D}}$).

\begin{figure}
	\centering
	\newcommand\width{1.0}
	\subfloat[Proposed UDA training setup]{
		\includegraphics[width=\width\linewidth]{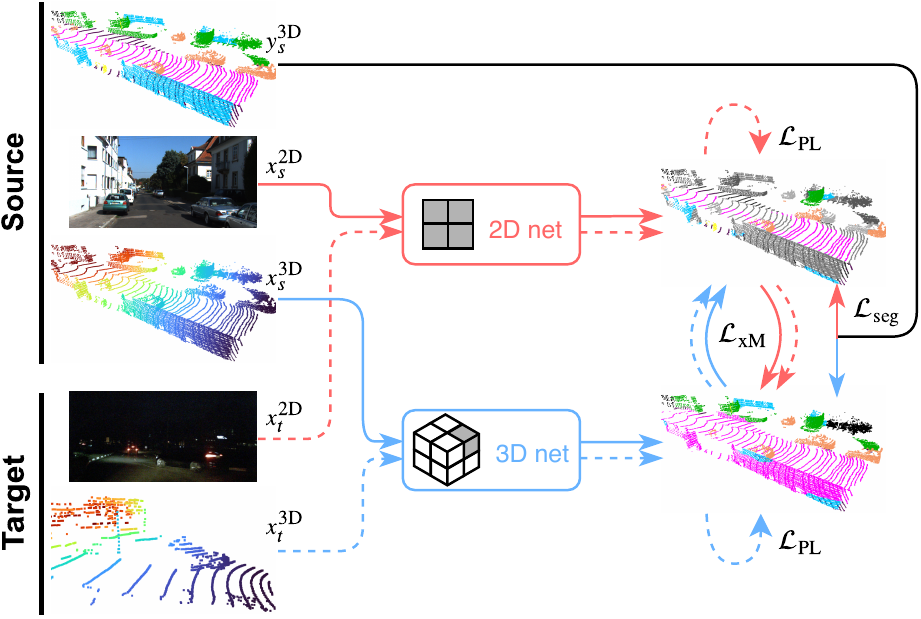}
		\label{fig:cml-uda_overview}
	}\\
	\subfloat[UDA on multi-modal data]{
		\includegraphics[width=\width\linewidth]{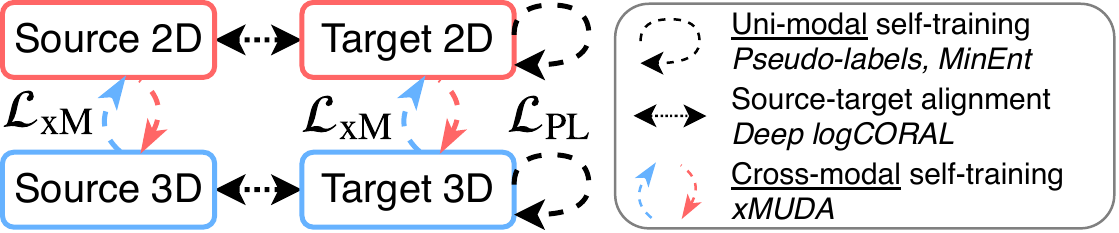}
		\label{fig:domainAdaptation}
	}
	\caption{\textbf{Details of proposed cross-modal training with adaptation}. \protect\subref{fig:cml-uda_overview} xMUDA learns from supervision on the source domain (plain lines) and self-supervision on the target domain (dashed lines), while benefiting from the cross-modal predictions of {\color{red}2D}/{\color{cyan}3D}.
	\protect\subref{fig:domainAdaptation} We consider four data subsets: Source 2D, Target 2D, Source 3D and Target 3D. In contrast to existing techniques, xMUDA introduces a cross-modal self-training mechanism for UDA.}
	\label{fig:uda}
	\vspace{-0.3cm}
\end{figure}

\subsection{Learning Scheme}\label{sec:CMLLearningScheme}

The goal of our cross-modal learning scheme is to exchange information between the modalities in a controlled manner to teach them to be aware of each other. 
This auxiliary objective can effectively improve the performance of each modality and does not require any annotations which enables its use for UDA on target dataset $\mathcal{T}$. 
In the following we define the basic supervised learning setup, our cross-modal loss $\mathcal{L}_{\text{xM}}$, and the additional pseudo-label learning method. The loss flows are depicted in Fig.\,\ref{fig:cml-uda_overview}.

\vspace{-0.2cm}
\paragraph{Supervised Learning.}

The main goal of 3D segmentation is learned through cross-entropy in a classical supervised fashion on the source data. We can write the segmentation loss $\mathcal{L}_\text{seg}$ for each network stream (2D and 3D) as: 
\begin{align}\label{eq:SegLoss}
\mathcal{L}_{\text{seg}}(\bm{x}_s, \ys) = - \frac{1}{N}\sum_{n=1}^N \sum_{c=1}^C \bm{y}_s^{(n, c)} \log \bm{P}_{\bm{x}_s}^{(n, c)},
\end{align}
where $\bm{x}_s$ is either $\xs{2D}$ or $\xs{3D}$.

\vspace{-0.2cm}
\paragraph{Cross-Modal Learning.}
The objective of unsupervised learning across modalities is twofold. Firstly, we want to transfer knowledge from one modality to the other on the target dataset. For example, let one modality be sensitive and the other more robust to the domain shift, then the robust modality should teach the sensitive modality the correct class in the target domain where no labels are available. Secondly, we want to design an auxiliary objective on source and target, where the task is to estimate the other modality's prediction. By mimicking not only the class with maximum probability, but the whole distribution, more information is exchanged, leading to softer labels.

We choose KL divergence for the cross-modal loss $\mathcal{L}_\text{xM}$ and define it as follows:
\begin{align}\label{eq:CMLLoss}
\mathcal{L}_{\text{xM}}(\bm{x}) &= \bm{D}_\text{KL}(\bm{P}_{\bm{x}}^{(n, c)} || \bm{Q}_{\bm{x}}^{(n, c)})\\
&= - \frac{1}{N} \sum_{n=1}^N \sum_{c=1}^C \bm{P}_{\bm{x}}^{(n, c)} \log \frac{\bm{P}_{\bm{x}}^{(n, c)}}{\bm{Q}_{\bm{x}}^{(n, c)}},
\end{align}
with $(\bm{P}, \bm{Q}) \in  \{(\bm{P}_\text{2D}, P_{\text{3D} \to \text{2D}}), (\bm{P}_\text{3D}, P_{\text{2D} \to \text{3D}})\}$ where $\bm{P}$ is the target distribution from the main prediction which is to be estimated by the mimicking prediction $\bm{Q}$.
This loss is applied on the source and the target domain as it does not require ground truth labels and is the key to our proposed domain adaptation framework. For source, $\mathcal{L}_\text{xM}$ can be seen as an auxiliary mimicry loss in addition to the main segmentation loss $\mathcal{L}_\text{seg}$.

The complete objective for each network stream (2D and 3D) is the combination of the segmentation loss $\mathcal{L}_\text{seg}$ on source and the cross-modal loss $\mathcal{L}_\text{xM}$ on source and target:
\begin{multline}\label{eq:completeObjective}
\min_{\theta}\Big[\frac{1}{|\mathcal{S}|}\sum_{\bm{x}_s\in\mathcal{S}} \Big(\mathcal{L}_\text{seg}(\bm{x}_s, \ys) + \lambda_s \mathcal{L}_\text{xM}(\bm{x}_s)\Big) \\
 + \frac{1}{|\mathcal{T}|}\sum_{\bm{x}_t\in\mathcal{T}} \lambda_t \mathcal{L}_\text{xM}(\bm{x}_t)\Big],
\end{multline}
where $\lambda_s, \lambda_t$ are hyperparameters to weight $\mathcal{L}_\text{xM}$ on source and target respectively and $\theta$ are the network weights of either the 2D or the 3D stream.

There are parallels between the cross-modal learning and model distillation which also adopts KL divergence as mimicry loss, but with the goal to transfer knowledge from a large network to a smaller one in a supervised setting~\cite{hinton2015distilling}.
Recently Zhang~\textit{et al.} introduced Deep Mutual Learning~\cite{zhang2018deep} where an ensemble of uni-modal networks are jointly trained to learn from each other in collaboration.
Though to some extent, our cross-modal learning is of similar nature to those strategies, we tackle a different distillation angle, i.e. across modalities (2D/3D) and not in the supervised, but in the UDA setting.

\vspace{-0.2cm}
\paragraph{Additional self-training with Pseudo-Labels.} Cross-modal learning is complementary to pseudo-labeling~\cite{lee2013pseudo} used originally in semi-supervised learning and recently in UDA ~\cite{li2019bidirectional,zou2019confidence}.
In details, once having optimized a model with Eq.\,\ref{eq:completeObjective}, we extract pseudo-labels offline, selecting highly confident labels based on the predicted class probability.
Then, we train again from scratch using the produced pseudo-labels for an additional segmentation loss on the target training set.
The optimization problem writes:
\begin{multline}\label{eq:completeObjectiveWithPL}
\min_{\theta}\Big[\frac{1}{|\mathcal{S}|}\sum_{\bm{x}_s} \Big(\mathcal{L}_\text{seg}(\bm{x}_s, \ys) + \lambda_s \mathcal{L}_\text{xM}(\bm{x}_s)\Big) \\
+ \frac{1}{|\mathcal{T}|}\sum_{\bm{x}_t} \Big(\lambda_\text{PL} \mathcal{L}_\text{seg}(\bm{x}_t, \hat{\bm{y}}^\text{3D}) + \lambda_t \mathcal{L}_\text{xM}(\bm{x}_t)\Big)\Big],
\end{multline}
where $\lambda_\text{PL}$ is weighting the pseudo-label segmentation loss and $\hat{\bm{y}}^\text{3D}$ are the pseudo-labels. For clarity, we will refer to the xMUDA variant that uses additional self-training with pseudo-labels as xMUDA\textsubscript{PL}.

\subsection{Discussion}
A central contribution of our work is the formulation of cross-modal learning via KL-divergence minimization in multi-modal scenarios which helps us not only to benefit from \emph{multiple sensors} but also to mitigate \emph{domain shift}.
Indeed, as computing KL-divergence between 2D and 3D predictions does not require ground-truth, our learning scheme allows extra regularization on the \emph{target} set -- bringing adaptation effects.
Fig.~\ref{fig:domainAdaptation} visualizes the~$4$ data subsets considered in our task and shows which dimension different UDA techniques operate on.
Opposed to previous UDA methods which only act on a single modality, xMUDA introduces a new way of cross-modal self-training and is thus orthogonal and complementary to existing adaptation techniques.

\begin{table*}
	\centering
	\footnotesize
	\begin{threeparttable}
        \begin{tabular}{lccccccccc}
        \toprule
        & \multicolumn{3}{c}{ USA/Singapore } & \multicolumn{3}{c}{ Day/Night } & \multicolumn{3}{c}{ A2D2/SemanticKITTI } \\
        \cmidrule(r){2-4}
        \cmidrule(r){5-7}
        \cmidrule(r){8-10}
        Method & 2D & 3D & softmax avg & 2D & 3D & softmax avg & 2D & 3D & softmax avg \\
        \midrule
        Baseline (source only) & 53.4 & 46.5 & 61.3 & 42.2 & 41.2 & 47.8 & 36.0 & 36.6 & 41.8 \\
        \midrule
        Deep logCORAL~\cite{morerio2017minimal} & 52.6 & 47.1 & 59.1 & 41.4 & 42.8 & \textbf{51.8} & 35.8\tnote{*} & 39.3 & 40.3 \\
        MinEnt~\cite{vu2019advent} & 53.4 & 47.0 & 59.7 & 44.9 & 43.5 & 51.3 & 38.8 & 38.0 & 42.7 \\
        PL~\cite{li2019bidirectional} & 55.5 & 51.8 & 61.5 & 43.7 & 45.1 & 48.6 & 37.4 & 44.8 & 47.7 \\
        \midrule
        xMUDA & 59.3 & 52.0 & 62.7 & 46.2 & 44.2 & 50.0 & 36.8 & 43.3 & 42.9 \\
        xMUDA\textsubscript{PL} & \textbf{61.1} & \textbf{54.1} & \textbf{63.2} & \textbf{47.1} & \textbf{46.7} & 50.8 & \textbf{43.7} & \textbf{48.5} & \textbf{49.1} \\
        \midrule
        Oracle & 66.4 & 63.8 & 71.6 & 48.6 & 47.1 & 55.2 & 58.3 & 71.0 & 73.7 \\
        \bottomrule
        \end{tabular}
	\begin{tablenotes}
		\item[*] Trained with batch size 6 instead of 8 to fit into GPU memory.
	\end{tablenotes}
	\end{threeparttable}
	\caption{mIoU on the respective target sets for 3D semantic segmentation in different cross-modal UDA scenarios. We report the result for each network stream (2D and 3D) as well as the ensembling result (`softmax avg').}
	\label{tab:resultsMain}
\end{table*}
\section{Experiments}
\subsection{Datasets}\label{sec:datasets}
To evaluate xMUDA, we identified 3 real-to-real adaptation scenarios.
In the \textbf{day-to-night} case, LiDAR has a small domain gap, as it is an active sensor sending out laser beams which are mostly invariant to lighting conditions. In contrast, camera has a large domain gap as its passive sensing suffers from lack of light sources, leading to drastic changes in object appearance.
The second scenario is \textbf{country-to-country} adaptation, where the domain gap can be larger for LiDAR or camera: for some classes the 3D shape might change more than the visual appearance or vice versa. The third scenario, \textbf{dataset-to-dataset}, comprises changes in the sensor setup, such as camera optics, but most importantly a higher LiDAR resolution on target. 3D networks are sensitive to varying point cloud density and the image could help to guide and stabilize adaptation.

We leverage recently published autonomous driving datasets nuScenes~\cite{nuscenes2019}, A2D2~\cite{aev2019} and SemanticKITTI~\cite{behley2019iccv} in which LiDAR and camera are synchronized and calibrated allowing to compute the projection between a 3D point and its corresponding 2D image pixel. The chosen datasets contain 3D annotations. For simplicity and consistency across datasets, we only use the front camera image and the LiDAR points that project into it.

For nuScenes, the annotations are 3D bounding boxes and we obtain the point-wise labels for 3D semantic segmentation by assigning the corresponding object label if a point lies inside a 3D box; otherwise the point is labeled as background. We use the meta data to generate the splits for two UDA scenarios: Day/Night and USA/Singapore.

A2D2 and SemanticKITTI provide segmentation labels. For UDA, we define 10 shared classes between the two datasets. The LiDAR setup is the main difference: in A2D2, there are 3 LiDARs with 16 layers which generate a rather sparse point cloud and in SemanticKITTI, there is one high-resolution LiDAR with 64 layers.

We provide the data split details in Appendix \ref{sec:datasetSplits}.

\subsection{Implementation Details}
\label{sec:implementation}

\paragraph{2D Network.}
We use a modified version of U-Net~\cite{ronneberger2015unet} with a ResNet34~\cite{he2016resnet} encoder where we add dropout after the 3rd and 4th layer and initialize with ImageNet pretrained weights provided by PyTorch. In the decoder, each layer consists of a transposed convolution, concatenation with encoder features of same resolution (skip connection) and another convolution to mix the features. The network takes an image $\bm{x}^\text{2D}$ as input and produces an output feature map with equal spatial dimensions $(H, W, F_\text{2D})$, where $F_\text{2D}$ is the number of feature channels. 
In order to lift the 2D features to 3D, we sample them at sparse pixel locations where the 3D points project into the feature map, and obtain the final two-dimensional feature matrix $(N, F_\text{2D})$.

\vspace{-0.2cm}
\paragraph{3D Network.}
For SparseConvNet~\cite{SparseConvNet} we leverage the official PyTorch implementation and a U-Net architecture with 6 times downsampling. We use a voxel size of 5cm which is small enough to only have one 3D point per voxel.

\vspace{-0.2cm}
\paragraph{Training.}
For data augmentation we employ horizontal flipping and color jitter in 2D, and x-axis flipping, scaling and rotation in 3D. Due to the wide angle image in SemanticKITTI, we crop a fixed size rectangle randomly on the horizontal image axis to reduce memory during training.
Log-smoothed class weights are used in all experiments to address class imbalance.
For the KL divergence for the cross-modal loss in PyTorch, we \textit{detach} the target variable to only backpropagate in either the 2D or the 3D network.
We use a batch size of 8, the Adam optimizer with $\beta_1=0.9, \beta_2=0.999$, and an iteration based learning schedule where the learning rate of $0.001$ is divided by 10 at 80k and 90k iterations; the training finishes at 100k.
We jointly train the 2D and 3D stream and at each iteration, accumulate gradients computed on source and target batch. All trainings fit into a single GPU with 11GB RAM.

For xMUDA, we train with Eq.\,\ref{eq:completeObjective}, where we apply the segmentation loss using ground truth labels on source and cross-modal loss on source \textit{and} target. For xMUDA\textsubscript{PL} , we generate pseudo-labels offline as in~\cite{li2019bidirectional} with the previously trained xMUDA model and train again from scratch, now with additional segmentation loss on target with pseudo-labels (Eq.\,\ref{eq:completeObjectiveWithPL}). Note, that we do not select the best weights on the validation set, but rather use the last checkpoint to generate the pseudo-labels in order to prevent any supervised learning signal. The 2D and 3D network are trained jointly and optimized on source and target at each iteration.

\subsection{Main Experiments}\label{sec:mainExperiments}

We evaluate our method on the 3 proposed cross-modal UDA scenarios and compare against uni-modal UDA methods: Deep logCORAL~\cite{morerio2017minimal}, entropy minimization (MinEnt)~\cite{vu2019advent} and pseudo-labeling (PL)~\cite{li2019bidirectional}.
Regarding PL, we apply~\cite{li2019bidirectional} as follows: we generate pseudo-labels offline with a first training without UDA, and discard unconfident labels through class-wise thresholding. Then, we run a second training from scratch adding PL loss on target. The image-2-image translation part was excluded due to its instability, high training complexity and incompatibility with LiDAR data, thus limiting reproducibility.
Regarding the two other uni-modal techniques, we adapt the published implementations to our settings.
For all, we searched for the best respective hyperparameters.

We report mean Intersection over Union (mIoU) results for 3D segmentation in Tab.\,\ref{tab:resultsMain} on the target test set for the 3 UDA scenarios. We evaluate on the test set using the checkpoint that achieved the best score on the validation set.
In addition to the scores of the 2D and 3D model, we show the ensembling result (`softmax avg') which is obtained by taking the mean of the predicted 2D and 3D probabilities after softmax.
The baseline is trained on source only and the oracle on target only, except the Day/Night oracle, where we used batches of 50\%/50\% Day/Night to prevent overfitting.
The uni-modal UDA baselines~\cite{li2019bidirectional,morerio2017minimal,vu2019advent} are applied separately on each modality.

xMUDA -- using the cross-modal loss but \textit{not} PL -- brings a significant adaptation effect on all 3 UDA scenarios compared to `Baseline (source only)' and often outperforms the uni-modal UDA baselines.
We observe, that xMUDA consistently improves \textit{both} modalities (2D and 3D), i.e. even the strong modality can learn from the weaker one.
xMUDA\textsubscript{PL} achieves the best score everywhere with the only exception of Day/Night softmax avg. Further, cross-modal learning and self-training with pseudo-labels (PL) are complementary as their combination in xMUDA\textsubscript{PL} consistently yields a higher score than each separate technique.

\begin{figure*}
	\centering
	\includegraphics[width=1\textwidth]{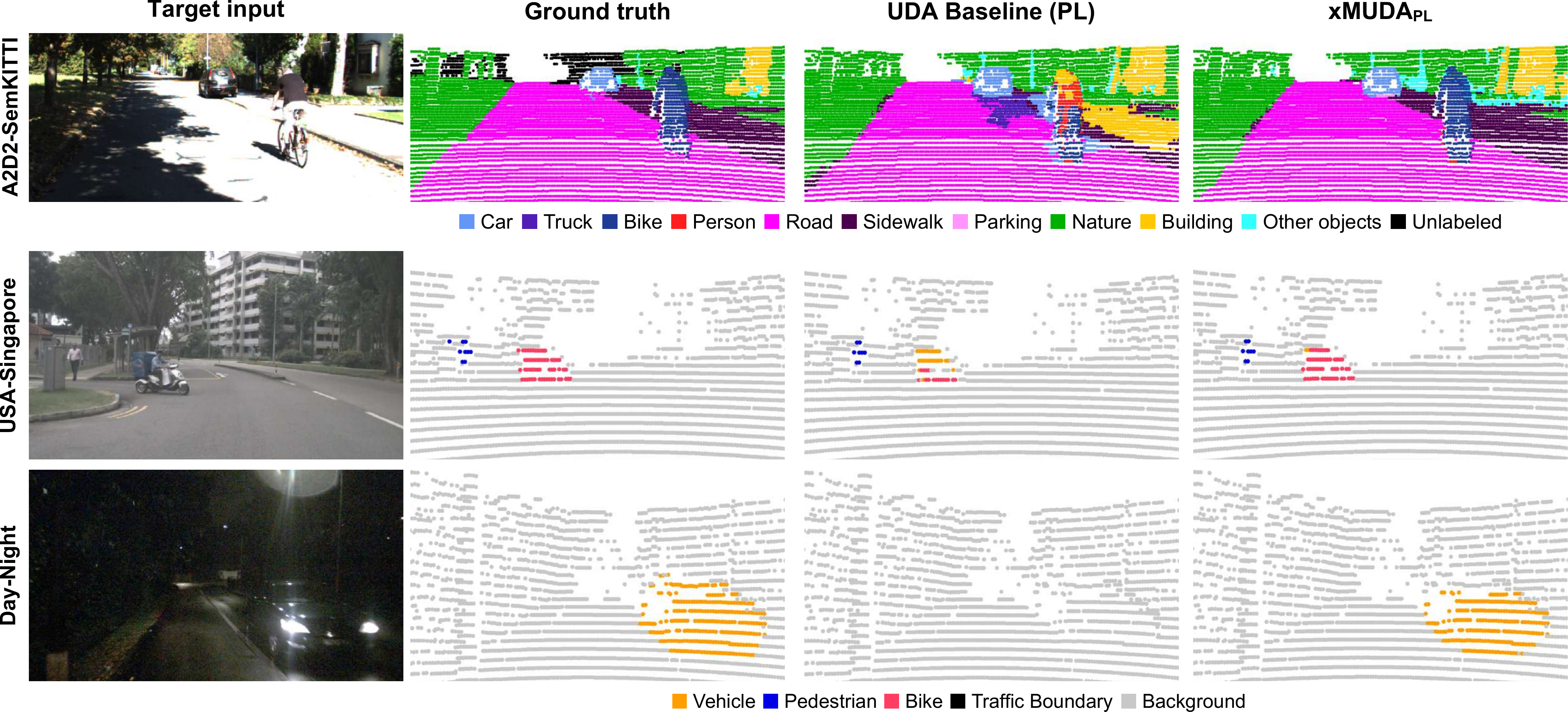}
	\caption{\textbf{Qualitative results on the 3 proposed splits}. We show the ensembling result obtained from averaging the softmax output of 2D and 3D on the UDA Baseline (PL) and xMUDA\textsubscript{PL}.\\
	\textbf{A2D2/SemanticKITTI}: xMUDA\textsubscript{PL} helps to stabilize and refine segmentation performance when there are sensor changes (3x16 layer LiDAR with different angles to 64 layer LiDAR).\\
	\textbf{USA/Singapore}: Delivery motorcycles with a storage box on the back are common in Singapore, but not in USA. The 3D shape might resemble a vehicle. However, 2D appearance information is leveraged in xMUDA\textsubscript{PL} to improve the recognition.\\
	\textbf{Day/Night}: The visual appearance of a car at night with headlights turned on is very different than during day. The uni-modal UDA baseline is not able to learn this new appearance. However, if information between camera and robust-at-night LiDAR is exchanged in xMUDA\textsubscript{PL}, it is possible to detect the car correctly at night.}
	\label{fig:qualitativeResults}
\end{figure*}

Qualitative results are presented in Fig.\,\ref{fig:qualitativeResults} and show the versatility of xMUDA across all proposed UDA scenarios. We provide additional qualitative results in Fig.~\ref{fig:qualitativeResultsSupp} and a video of the A2D2 to SemanticKITTI scenario at \url{http://tiny.cc/xmuda}.

\subsection{Extension to Fusion}\label{sec:extFusion}
\begin{figure}
	\centering
	\newcommand\height{0.45}
	\subfloat[Vanilla Fusion]{
		\includegraphics[height=\height\linewidth]{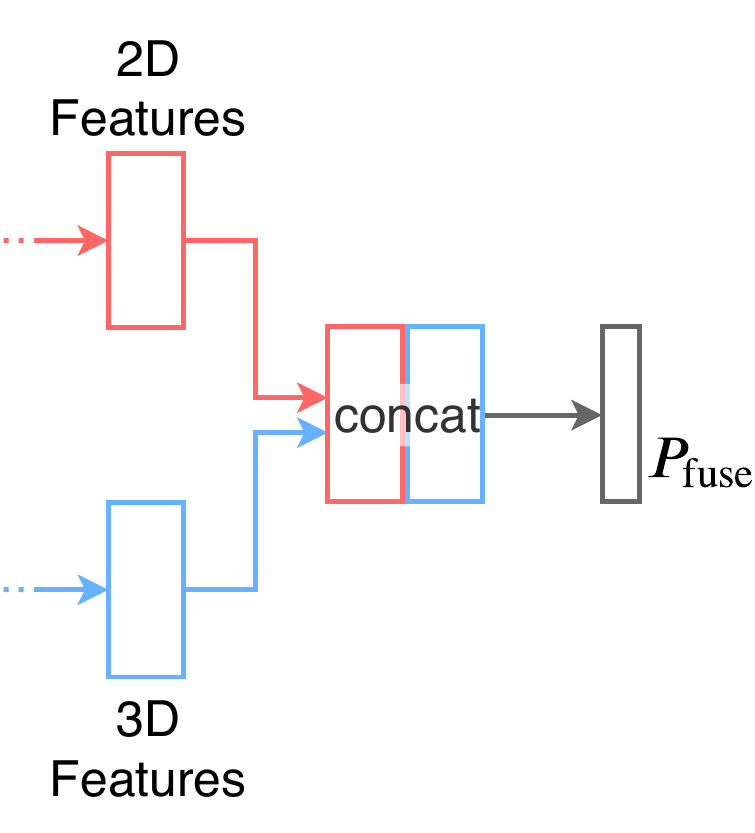}
		\label{fig:architectureFusionVanilla}
	}
	\subfloat[xMUDA Fusion]{
		\includegraphics[height=\height\linewidth]{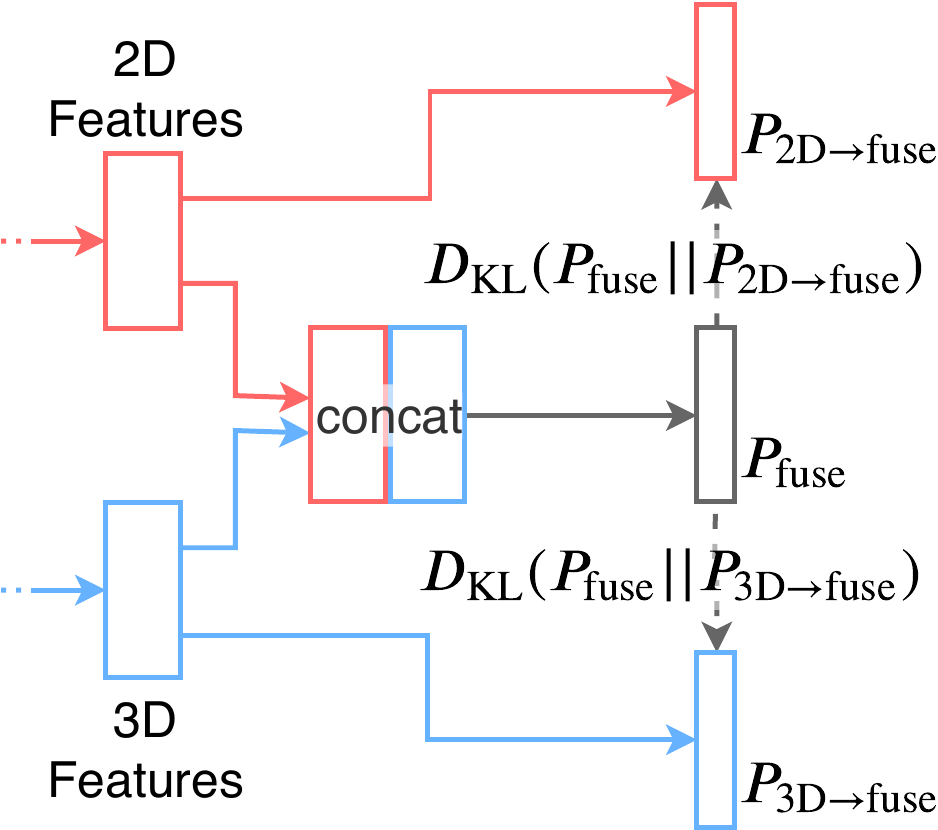}
		\label{fig:architectureFusionAS}
	}
	\caption{\textbf{Architectures for fusion}. \protect\subref{fig:architectureFusionVanilla} In Vanilla Fusion the 2D and 3D features are concatenated, fed into a linear layer with ReLU to mix the features and followed by another linear layer and softmax to obtain a fused prediction $\boldsymbol{P}_\text{fuse}$. \protect\subref{fig:architectureFusionAS} In xMUDA Fusion, we add two uni-modal outputs $\boldsymbol{P}_{\text{2D} \to \text{fuse}}$ and $\boldsymbol{P}_{\text{3D} \to \text{fuse}}$ that are used to mimic the fusion output $\boldsymbol{P}_\text{fuse}$.}
	\label{fig:architectureVanillaVsAS}
	\vspace{-0.3cm}
\end{figure}
In Sec.\,\ref{sec:mainExperiments} we show how each 2D and 3D modality can be improved with xMUDA.
However, can we obtain even better results with \textit{fusion}?

\begin{figure}[h!]
    \vspace{-0.3cm}
	\centering
	\subfloat[Single head architecture]{
		\includegraphics[height=0.42\linewidth]{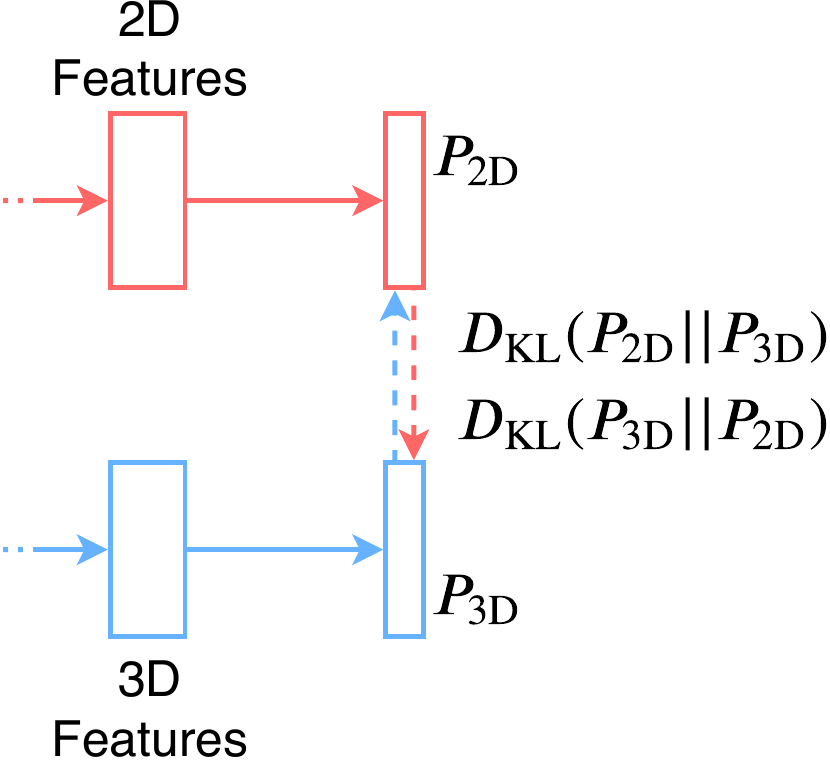}
		\label{fig:architectureVanilla}
	}
	\hfill
	\subfloat[$\lambda_s=1.0$, varying $\lambda_t$]{
		\includegraphics[height=0.4\linewidth]{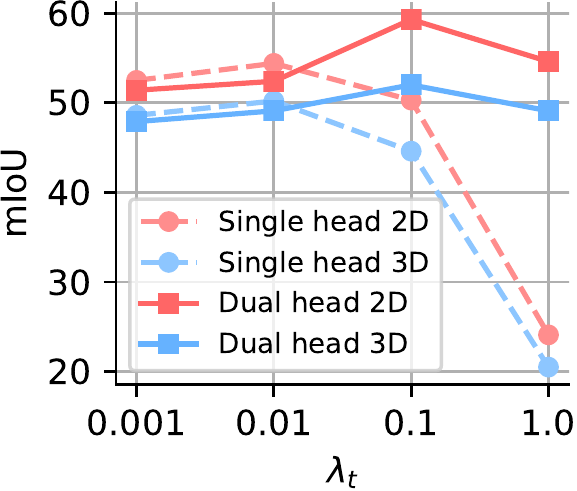}
		\label{fig:hyperparam}
	}
    \vspace{-0.1cm}
	\caption{\textbf{Single vs. Dual segmentation head.} \protect\subref{fig:architectureVanilla} Main and mimicry prediction are not uncoupled as in xMUDA of Fig.\,\ref{fig:architecture}.
    \protect\subref{fig:hyperparam}
    Curves of mIoU w.r.t. $\lambda_t$ of single- \textit{vs.} dual-head architectures.
    USA/Singapore scenario.
    }
	\label{fig:architectureVanillaVsAS}
	\vspace{-0.3cm}
\end{figure}

A common fusion architecture is late fusion where the features from different sources are concatenated (see Fig.\,\ref{fig:architectureFusionVanilla}). However, when merging the main 2D/3D branches into a unique fused head, we can no longer apply cross-modal learning (as in Fig.\,\ref{fig:architectureVanilla}). To address this problem, we propose `xMUDA Fusion' where we add an additional segmentation head to both 2D and 3D network streams \textit{prior} to the fusion layer, with the purpose of mimicking the central fusion head (see Fig.\,\ref{fig:architectureFusionAS}). Note that this idea could also be applied on top of other fusion architectures.

\begin{table}
	\scriptsize
	\centering
    \begin{tabular}{llc}
        \toprule
        & Architecture & mIoU \\
        \midrule
        Baseline (source only) & Vanilla & 59.9 \\
        \midrule
        Deep logCORAL~\cite{morerio2017minimal} & Vanilla & 58.2 \\
        MinEnt~\cite{vu2019advent} & Vanilla & 60.8 \\
        PL~\cite{li2019bidirectional} & Vanilla & 65.2 \\
        Distillation & Vanilla & 65.8 \\
        \midrule
        xMUDA Fusion & xMUDA & 61.9 \\
        xMUDA\textsubscript{PL} Fusion & xMUDA & \textbf{66.6} \\
        \midrule
        Oracle & xMUDA & 72.2 \\
        \bottomrule
    \end{tabular}
	\caption{mIoU for fusion methods, USA/Singapore scenario. In `Distillation' we use the xMUDA\textsubscript{PL} model of the main experiments reported in Tab.\,\ref{tab:resultsMain} to generate pseudo-labels from the softmax average and use those to train the Vanilla Fusion network.}
	\label{tab:fusion}
	\vspace{-0.3cm}
\end{table}

In Tab.\,\ref{tab:fusion} we show results for different fusion approaches where we specify which architecture was used (Vanilla late fusion from Fig.\,\ref{fig:architectureFusionVanilla} or xMUDA Fusion from Fig.\,\ref{fig:architectureFusionAS}).
While `xMUDA\textsubscript{PL} Fusion' outperforms all other uni-modal baselines, `xMUDA Fusion' already achieves better performances than `Deep logCORAL' and `MinEnt'.

\section{Ablation Studies}

\subsection{Segmentation Heads}\label{sec:segmentationHeads}

In the following we justify our design choice of two segmentation heads per modality stream as opposed to a single one in a naive approach (see Fig.\,\ref{fig:architectureVanilla}).
In the single head architecture the mimicking objective is directly applied between the 2 main predictions. There is shared information between 2D/3D, but also private information in each modality. An unwanted solution to reduce the cross-modal loss $\mathcal{L}_{\text{xM}}$ is that the networks discard private information, so that they both only use shared information, making it easier to align their outputs. However, we conjecture that best performance can be achieved if private information is also used. By separating the main from the mimicking prediction with dual segmentation heads, we can effectively decouple the two optimization objectives: The main head outputs the best possible prediction to optimize the segmentation loss, while the mimicking head can align with the other modality.

To benchmark single vs.~dual segmentation head architectures, we apply cross-modal loss $\mathcal{L}_{\text{xM}}$ only, excluding PL as it does not depend on the dual head approach.
We fix $\mathcal{L}_{\text{xM}}$ loss weight on source to $\lambda_s = 1.0$ and vary target $\lambda_t$. The hyperparameter $\lambda_t$ is at the focus of this analysis because it controls modality alignment on the target set, the main driver for UDA. In Fig.\,\ref{fig:hyperparam}, we show that best performance is achieved with the dual head architecture of xMUDA, while the single head architecture drops drastically in performance for high $\lambda_t$. We hypothesize that dual head is more robust because it disentangles the segmentation from the mimicking objective.

\subsection{Cross-Modal Learning on Source}
In (\ref{eq:completeObjective}), cross-modal loss $\mathcal{L}_{\text{xM}}$ is applied on source \textit{and} target, although we already have supervised segmentation loss $\mathcal{L}_{\text{seg}}$ on source. We observe a gain of 4.8 mIoU on 2D and 4.4 on 3D when adding $\mathcal{L}_{\text{xM}}$ on source as opposed to applying it on target \textit{only}. This shows that it is important to train the mimicking head on source, stabilizing the predictions, which can be exploited during adaptation on target.

\subsection{Cross-Modal Learning for Oracle Training}
We have shown that cross-modal learning is very effective for UDA. However, it can also be used in a purely supervised setting. When training the oracle with cross-modal loss $\mathcal{L}_{\text{xM}}$, we can improve over the baseline, see Tab.\,\ref{tab:supervisedResults}. We conjecture that $\mathcal{L}_{\text{xM}}$ is a beneficial auxiliary loss and can help to regularize training and prevent overfitting.

\begin{table}[h]
	\scriptsize
	\centering
    \begin{tabular}{lccc|cc}
        \toprule
        Method & 2D & 3D & softmax avg & Method & fusion \\
        \midrule
        w/o $\mathcal{L}_{\text{xM}}$ & 65.8 & 63.2 & 71.1 & Vanilla Fusion & 71.0 \\
        with $\mathcal{L}_{\text{xM}}$ & \textbf{66.4} & \textbf{63.8} & \textbf{71.6} & Fusion + $\mathcal{L}_{\text{xM}}$ & \textbf{72.2} \\
        \bottomrule
    \end{tabular}
\caption{Cross-modal loss in supervised setting for oracle training. mIoU on USA/Singapore.}
	\label{tab:supervisedResults}
\end{table}

\vspace{-0.2em}
\section{Conclusion}

We propose xMUDA, Cross-Modal Unsupervised Domain Adaptation, where modalities learn from each other to improve performance on the target domain. For cross-modal learning we introduce mutual mimicking between the modalities, achieved through KL divergence. We design an architecture with separate main and mimicking head to disentangle the segmentation from the cross-modal learning objective. Experiments on 3D semantic segmentation on new UDA scenarios using 2D/3D datasets, show that xMUDA largely outperforms uni-modal UDA and is complementary to the pseudo-label strategy. An analog performance boost is observed on fusion.

We think that cross-modal learning could be useful in a wide variety of settings and tasks, not limited to UDA. Particularly, it should be beneficial for supervised learning and other modalities than image and point cloud.

\appendix
\section{Dataset Splits}\label{sec:datasetSplits}
\subsection{nuScenes}
The nuScenes dataset~\cite{nuscenes2019} consists of 1000 driving scenes, each of 20 seconds, which corresponds to 40k annotated keyframes taken at 2Hz. The scenes are split into train (28,130 keyframes), validation (6,019 keyframes) and hidden test set. The point-wise 3D semantic labels are obtained from 3D boxes like in~\cite{wu2018squeezeseg}. We propose the following splits destined for domain adaptation with the respective source/target domains: Day/Night and Boston/Singapore. Therefore, we use the official validation split as test set and divide the training set into train/val for the target set (see Tab.~\ref{tab:splits} for the number of frames in each split). As the number of object instances in the target split can be very small (e.g. for night), we merge the objects into 5 categories: \textbf{vehicle} (car, truck, bus, trailer, construction vehicle), \textbf{pedestrian}, \textbf{bike} (motorcycle, bicycle), \textbf{traffic boundary} (traffic cone, barrier) and \textbf{background}.

\begin{table}
	\scriptsize
	\centering
	\begin{tabular}{lrrrrr}
	    \toprule
        & \multicolumn{2}{c}{ source } & \multicolumn{3}{c}{ target } \\
        \cmidrule(r){2-3}
        \cmidrule(r){4-6}
        Split & train & test & train & val & test \\
        \midrule
        Day - Night & 24,745 & 5,417 & 2,779 & 606 & 602 \\
        Boston - Singapore & 15,695 & 3,090 & 9,665 & 2,770 & 2,929 \\
        A2D2 - SemanticKITTI & 27,695 & 942 & 18,029 & 1,101 & 4,071 \\
        \bottomrule
    \end{tabular}
	\caption{Number of frames for the 3 splits.}
	\label{tab:splits}
\end{table}

\subsection{A2D2 and SemanticKITTI}
The A2D2 dataset \cite{aev2019} features 20 drives, which corresponds to 28,637 frames. The point cloud comes from three 16-layer front LiDARs (left, center, right) where the left and right front LiDARS are inclined. The semantic labeling was carried out in the 2D image for 38 classes and we compute the 3D labels by projection of the point cloud into the labeled image. We keep scene 20180807\_145028 as test set and use the rest for training.

The SemanticKITTI dataset \cite{behley2019iccv} provides 3D point cloud labels for the Odometry dataset of Kitti~\cite{geiger2012cvpr} which features large angle front camera and a 64-layer LiDAR. The annotation of the 28 classes has been carried out directly in 3D. We use the scenes $\{0, 1, 2, 3, 4, 5, 6, 9, 10\}$ as train set, $7$ as validation and $8$ as test set.

We select 10 shared classes between the 2 datasets by merging or ignoring them (see Tab.~\ref{tab:classMapping}). The 10 final classes are car, truck, bike, person, road, parking, sidewalk, building, nature, other-objects.

{\small
\bibliographystyle{ieee_fullname}
\bibliography{egbib}
}

\begin{table*}
	\scriptsize
	\centering
	\begin{tabular}{ll|ll}
    A2D2 class & mapped class & SemanticKITTI class & mapped class \\
    \toprule
    Car 1 & car & unlabeled & ignore \\
    Car 2 & car & outlier & ignore \\
    Car 3 & car & car & car \\
    Car 4 & car & bicycle & bike \\
    Bicycle 1 & bike & bus & ignore \\
    Bicycle 2 & bike & motorcycle & bike \\
    Bicycle 3 & bike & on-rails & ignore \\
    Bicycle 4 & bike & truck & truck \\
    Pedestrian 1 & person & other-vehicle & ignore \\
    Pedestrian 2 & person & person & person \\
    Pedestrian 3 & person & bicyclist & bike \\
    Truck 1 & truck & motorcyclist & bike \\
    Truck 2 & truck & road & road \\
    Truck 3 & truck & parking & parking \\
    Small vehicles 1 & bike & sidewalk & sidewalk \\
    Small vehicles 2 & bike & other-ground & ignore \\
    Small vehicles 3 & bike & building & building \\
    Traffic signal 1 & other-objects & fence & other-objects \\
    Traffic signal 2 & other-objects & other-structure & ignore \\
    Traffic signal 3 & other-objects & lane-marking & road \\
    Traffic sign 1 & other-objects & vegetation & nature \\
    Traffic sign 2 & other-objects & trunk & nature \\
    Traffic sign 3 & other-objects & terrain & nature \\
    Utility vehicle 1 & ignore & pole & other-objects \\
    Utility vehicle 2 & ignore & traffic-sign & other-objects \\
    Sidebars & other-objects & other-object & other-objects \\
    Speed bumper & other-objects & moving-car & car \\
    Curbstone & sidewalk & moving-bicyclist & bike \\
    Solid line & road & moving-person & person \\
    Irrelevant signs & other-objects & moving-motorcyclist & bike \\
    Road blocks & other-objects & moving-on-rails & ignore \\
    Tractor & ignore & moving-bus & ignore \\
    Non-drivable street & ignore & moving-truck & truck \\
    Zebra crossing & road & moving-other-vehicle & ignore \\
    Obstacles / trash & other-objects & & \\
    Poles & other-objects & & \\
    RD restricted area & road & & \\
    Animals & other-objects & & \\
    Grid structure & other-objects & & \\
    Signal corpus & other-objects & & \\
    Drivable cobbleston & road & & \\
    Electronic traffic & other-objects & & \\
    Slow drive area & road & & \\
    Nature object & nature & & \\
    Parking area & parking & & \\
    Sidewalk & sidewalk & & \\
    Ego car & car & & \\
    Painted driv. instr. & road & & \\
    Traffic guide obj. & other-objects & & \\
    Dashed line & road & & \\
    RD normal street & road & & \\
    Sky & ignore & & \\
    Buildings & building & & \\
    Blurred area & ignore & & \\
    Rain dirt & ignore & & \\
    \bottomrule
\end{tabular}
	\caption{Class mapping for A2D2 - SemanticKITTI UDA scenario.}
	\label{tab:classMapping}
\end{table*}

\begin{figure*}
	\centering
	\includegraphics[width=1\textwidth]{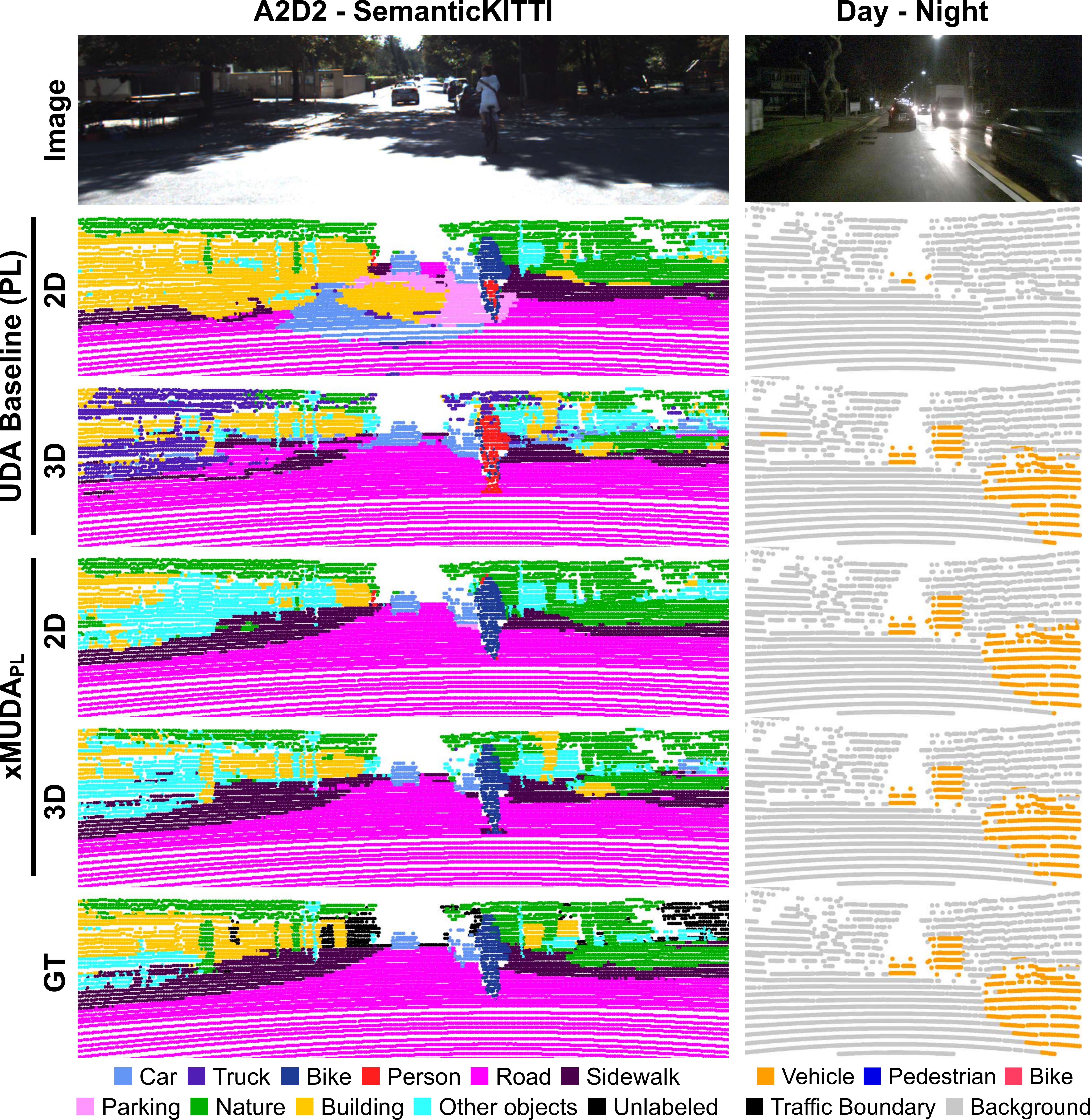}
	\caption{\textbf{Qualitative results on two UDA scenarios}. For UDA Baseline (PL) and xMUDA\textsubscript{PL}, we separately show the predictions of the 2D and 3D network stream.\\
	\textbf{A2D2/SemanticKITTI}: For the uni-modal UDA baseline (PL), the 2D prediction lacks consistency on the road and 3D is unable to recognize the bike and the building on the left correctly. In xMUDA\textsubscript{PL}, both modalities can stabilize each other and obtain better performance on the bike, the road, the sidewalk and the building. \\
	\textbf{Day/Night}: For the UDA Baseline, 2D can only partly recognize one car out of three while the 3D prediction is almost correct, with one false positive car on the left. With xMUDA\textsubscript{PL}, the 2D and 3D predictions are both correct.}
	\label{fig:qualitativeResultsSupp}
\end{figure*}

\end{document}